\pdfoutput=1
\documentclass[11pt]{article}

\usepackage[table,xcdraw]{xcolor}
\usepackage[final]{acl}

\usepackage{times}
\usepackage{latexsym}

\usepackage[T1]{fontenc}
\usepackage[utf8]{inputenc}

\usepackage{microtype}

\usepackage{inconsolata}

\usepackage{graphicx}

\usepackage{amsmath}
\usepackage{amssymb}
\usepackage[ruled,vlined]{algorithm2e}
\definecolor{mypink1}{rgb}{0.858, 0.188, 0.478}
\definecolor{codegreen}{rgb}{0,0.6,0}
\definecolor{codegray}{rgb}{0.5,0.5,0.5}
\definecolor{codepurple}{rgb}{0.58,0,0.82}
\definecolor{backcolour}{rgb}{0.95,0.95,0.92}
\usepackage{listings}
\usepackage{multirow}
\usepackage[normalem]{ulem}
\usepackage{arydshln}
\usepackage{booktabs}

\useunder{\uline}{\ul}{}

\title{Stepwise Reasoning Checkpoint Analysis:\\A Test Time Scaling Method to Enhance LLMs' Reasoning}
\author{
  Zezhong Wang$^{1}$\thanks{Work done during internship at Huawei Noah’s Ark Lab.}, Xingshan Zeng$^{2}$\thanks{Corresponding author}, Weiwen Liu$^{3}$, Yufei Wang$^{2}$, Liangyou Li$^{2}$, \\ 
  \bf Yasheng Wang$^{2}$, Lifeng Shang$^{2}$, Xin Jiang$^{2}$, Qun Liu$^{2}$, Kam-Fai Wong$^{1}$ \\
  $^1$The Chinese University of Hong Kong\\
  $^2$Huawei Noah’s Ark Lab, $^3$Shanghai Jiao Tong University\\
  \texttt{zzwang@se.cuhk.edu.hk, zeng.xingshan@huawei.com}
  }

\begin{document}
\maketitle
\begin{abstract}
    Mathematical reasoning through Chain-of-Thought (CoT) has emerged as a powerful capability of Large Language Models (LLMs), which can be further enhanced through Test-Time Scaling (TTS) methods like Beam Search and DVTS. However, these methods, despite improving accuracy by allocating more computational resources during inference, often suffer from path homogenization and inefficient use of intermediate results.
    To address these limitations, we propose Stepwise Reasoning Checkpoint Analysis (SRCA), a framework that introduces checkpoints between reasoning steps. 
    It incorporates two key strategies: (1) Answer-Clustered Search, which groups reasoning paths by their intermediate checkpoint answers to maintain diversity while ensuring quality, and (2) Checkpoint Candidate Augmentation, which leverages all intermediate answers for final decision-making. Our approach effectively reduces path homogenization and creates a fault-tolerant mechanism by utilizing high-quality intermediate results. Experimental results show that SRCA improves reasoning accuracy compared to existing TTS methods across various mathematical datasets.
\end{abstract}

\section{Introduction}\vspace{-0.1cm}
Large Language Models (LLMs) have demonstrated mathematical reasoning capabilities through Chain-of-Thought (CoT)~\cite{10.5555/3600270.3602070}. Recent studies indicate that Test Time Scaling (TTS), which expands test-time computing resources to allocate more reasoning budget through methods such as Beam Search~\cite{snell2024scalingllmtesttimecompute} and Diverse Verifier Tree Search (DVTS)~\cite{beeching2024scalingtesttimecompute}, can significantly improve accuracy in mathematical reasoning tasks~\cite{ji2025testtimecomputesystem1thinking,zhao2024marcoo1openreasoningmodels,chen2025simpleprovablescalinglaws}. These methods allow LLMs to sample multiple candidates at each reasoning step and score them using a process reward model (PRM)~\cite{xi2024enhancingllmreasoningcritique, wu2024exampleshighlevelautomatedreasoning, wang-etal-2024-math,zhang2025generativeverifiersrewardmodeling}. According to their strategies, they select high-scoring steps to continue reasoning, thus overcoming the limitations of single-path reasoning.

However, current methods face two key challenges in practice. First, maintaining diversity in the sampled reasoning paths is both crucial and arduous~\cite{misaki2025widerdeeperscalingllm,li-etal-2023-making}. Even though the model generates multiple candidate paths, the chosen ones usually follow similar reasoning directions. This happens because the reward mechanism favors local optimal solutions, causing the search process to converge too early and fail to explore diverse reasoning patterns~\cite{hooper2025etsefficienttreesearch, zeng2025bstarmonitoringbalancingexploration}. Second, existing methods underutilize intermediate reasoning results: many intermediate branches are discarded during the search, and only a few complete paths are used in the final decision, leading to a waste of computational resources~\cite{wang2024selfimprovementllmsmctsleveraging, zhang2025whathowwherewell}.

\begin{figure*}[ht]
    \centering
    \includegraphics[width=0.9\linewidth]{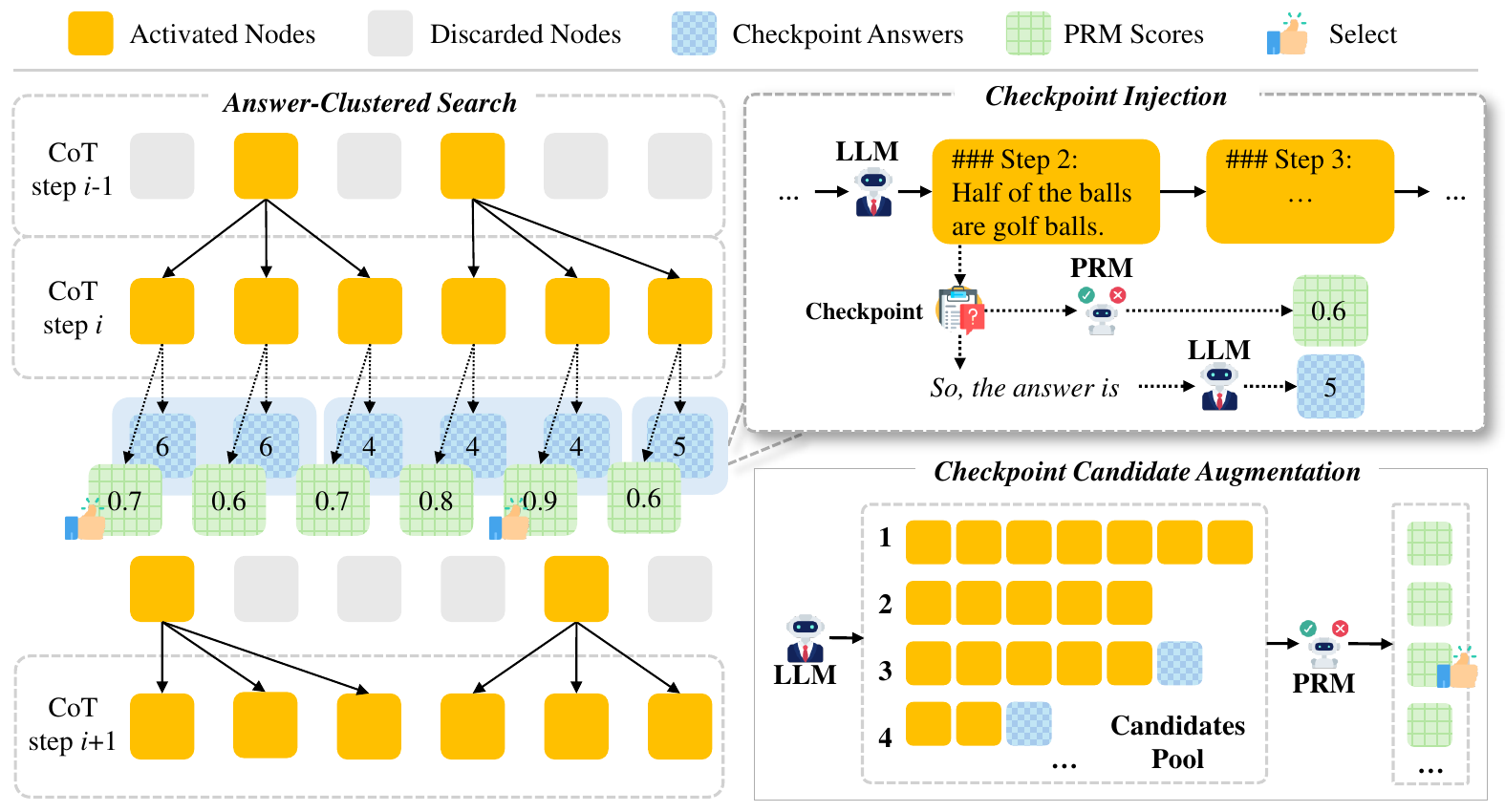}
    \caption{Overview of SRCA. 
    Top-right: The checkpoint operation, which serves as the atomic operation in SRCA. 
    Left: Illustration of ACS strategy at step $i$, where $N=6$ and $M=2$. Retrieved reasoning steps are clustered into three groups based on their checkpoint answers (indicated by different shades), with the highest-scoring nodes selected from clusters with answers 6 and 4 for subsequent reasoning.
    Bottom-right: CCA strategy, where paths 3 and 4 represent high-quality intermediate reasoning steps collected by CCA.
    }
    \label{fig:pipeline}
    \vspace{-0.3cm}
\end{figure*}

To address these issues, we propose \textbf{Stepwise Reasoning Checkpoint Analysis (SRCA)}. We introduce reasoning checkpoints as a foundational technique and propose a searching method and a decision-enhancement strategy based on it. 
We inject "checkpoints" after each reasoning step. Specifically, once a step is completed, we temporarily interrupt the reasoning process and append the prompt "\textit{So the answer is}" to the current reasoning steps, compelling the model to generate an intermediate prediction rather than continuing its reasoning process, as illustrated in the upper right corner of Figure~\ref{fig:pipeline}.
Using the intermediate answers collected at these checkpoints, we further propose an \textbf{Answer-Clustered Search} strategy. We group multiple reasoning steps sampled at the current checkpoint according to their detected answers, and retain high-quality reasoning steps from each group for further extension. 
This approach allows us to maintain multiple potential reasoning paths leading to different answers, thus increasing the diversity of reasoning processes and mitigating the issue of path homogenization.
Additionally, we introduce the \textbf{Checkpoint Candidate Augmentation} strategy. By collecting all intermediate answers generated at checkpoints, we expand the pool of candidate reasoning paths, allowing these intermediate results to participate in the final decision-making process. In this way, we fully utilize all high-quality intermediate results generated during reasoning, creating a fault-tolerant mechanism. Even if subsequent reasoning deviates, the retained high-quality intermediate predictions may still lead to the correct answer.

The contributions of this work can be summarized as follows:\vspace{-0.2cm}

\begin{itemize}
    \item We introduce the concept of reasoning checkpoints, providing a new methodology for analyzing and improving LLM reasoning processes during test time.\vspace{-0.2cm}
    \item Based on this concept, we develop SRCA, a framework that effectively addresses both path diversity and computational efficiency challenges.\vspace{-0.2cm}
    \item We conduct extensive experiments that demonstrate the superiority of our approach and provide valuable insights for future research in Test-Time Scaling.\vspace{-0.2cm}
\end{itemize}
\section{Related Works}

As enthusiasm for scaling pre-training computation wanes, Test-Time Scaling (TTS) has emerged as a key research focus~\cite{wang2024openropensourceframework, wu2024comparativestudyreasoningpatterns, chen2025simpleprovablescalinglaws}. TTS allocates additional computation during inference to improve performance, significantly enhancing LLMs' problem-solving capabilities across specialized and general tasks. Some TTS approaches use training to encourage LLMs to generate more extensive outputs for deeper reasoning~\cite{guan2025rstar, xi2024enhancingllmreasoningcritique}. These methods create synthetic data, including long chain-of-thought~\cite{chen2025empiricalstudyelicitingimproving, xiang20252reasoningllmslearning} and reflection-based examples~\cite{bi2025forestofthoughtscalingtesttimecompute, zhang2024accessinggpt4levelmathematical, yu2024improving}, to fine-tune LLMs, shifting their behavior from rapid responses to more deliberate reasoning.

Another category is training-free tree search, which forms the primary focus of this work~\cite{luo2024improvemathematicalreasoninglanguage, 10.5555/3692070.3694110, guan2024searchverifyfeedbackgeneration}. These methods dynamically guide the LLM's reasoning process using external verifiers or PRMs~\cite{jiang2024enhancingllmreasoningrewardguided, uesato2022solvingmathwordproblems, setlur2024rewardingprogressscalingautomated}. \citet{snell2024scalingllmtesttimecompute} introduced Beam Search to explore the reasoning space, where PRM evaluates each reasoning step and maintains a fixed number of promising paths based on the beam width. Building upon this foundation, subsequent research~\cite{beeching2024scalingtesttimecompute} proposed Diverse Verifier Tree Search (DVTS), which offers a notable improvement~\cite{liu20251bllmsurpass405b}. Instead of maintaining a single search beam, DVTS operates multiple search trees simultaneously, selecting and expanding the most promising reasoning path within each tree. Tree search algorithms, however, face two crucial challenges: the diversity problem and the utilization problem.

The diversity problem arises when PRMs inadvertently suppress the LLM's sampling diversity~\cite{chen2025empiricalstudyelicitingimproving, zheng2024processbenchidentifyingprocesserrors}. This occurs because only high-scoring paths are retained, and these paths often share similar problem-solving approaches. This issue is further exacerbated by the inherent biases in the imperfect PRMs or verifier~\cite{he2025largelanguagemodelsdetect, zheng2024processbenchidentifyingprocesserrors}. The utilization problem arises because tree search algorithms explore numerous paths, but typically only one contributes to the final result. This leads to many branches and intermediate processes being discarded, with utilization efficiency decreasing as search scale increases. The challenge is to efficiently integrate generated reasoning overhead~\cite{wang2024selfimprovementllmsmctsleveraging, zhang2025whathowwherewell, sui2025stopoverthinkingsurveyefficient}. This issue has evolved into the "over-thinking problem," where LLMs waste resources on simple problems, potentially leading to performance degradation through error accumulation~\cite{li2024escapeskyhighcostearlystopping, wu2025lessunderstandingchainofthoughtlength, huang2025efficienttesttimescalingselfcalibration, gan2025rethinkingexternalslowthinkingsnowball,aggarwal2025l1controllinglongreasoning}.

To address these dual challenges, we propose two novel strategies. First, the Answer Cluster Search algorithm is designed to enhance the diversity of tree search processes. Second, we introduce the Check Point Augmentation strategy to preserve high-quality intermediate reasoning processes for reuse in final answer decision-making, thereby addressing the low utilization problem inherent in tree search methodologies.

\section{Methodology}

In this section, we introduce three key techniques, Checkpoint Injection, which serves as the atomic operation in SRCA, Answer-Clustered Search (ACS), and Checkpoint Candidate Augmentation (CCA), to improve LLM's reasoning path searching.

\subsection{Checkpoint Injection}

We introduce a dynamic intervention mechanism to analyze the model’s reasoning trajectory through checkpoint injection. As shown in the upper top-right part of Figure~\ref{fig:pipeline}, the core procedure begins by monitoring the model’s output stream for predefined step delimiter tokens (e.g., \textit{"\#\#\# Step"}), which indicate the completion of a logical reasoning unit. 
Upon detecting such tokens, we inject a checkpoint to temporarily suspend autoregressive generation. At each checkpoint position, a fixed prompt template $x_{\text{ckpt}}=$ \textit{"So, the answer is "} is inserted to force the model to generate an intermediate prediction based solely on the accumulated context up to that step. The model’s immediate response to $x_{\text{ckpt}}$ is recorded as a checkpoint answer $a_t$ at step $t$, after which the LLM rolls back the generation state to the original checkpoint position. This rollback operation ensures the elimination of checkpoint influence from the ongoing reasoning process while preserving the model’s KV cache for continued generation~\cite{wang2025chainofprobeexaminingnecessityaccuracy}. The checkpoint answers subsequently serve as crucial criteria for path similarity assessment and grouping in the ACS strategy, while also being collected by the CCA method to enrich the final answer candidate pool.

\subsection{Answer-Clustered Search}
Similar to Beam Search~\cite{snell2024scalingllmtesttimecompute}, The Answer-Clustered Search (ACS) evaluates and retains a select few of the multiple reasoning steps sampled by the LLM for further reasoning. It enhances reasoning diversity through stepwise answer-guided clustering. In the following part, we will detail the four key steps of ACS using the running case shown on the left side of Figure~\ref{fig:pipeline}. 

\textbf{1. Sampling.} At each reasoning step $t$, we first determine the branching factor: for the initial step ($t=1$), the LLM samples $N$ candidate reasoning paths; for subsequent steps, each of the $M$ surviving beams generates $N/M$ sub-paths, maintaining a total budget of $N$ paths. This set of paths is defined as $\{p_t^{(j)}\}_{j=1}^N$. Figure~\ref{fig:pipeline} illustrates the case where $N=6$ and $M=2$.

\textbf{2. Clustering.} All $N$ paths undergo Checkpoint Injection at step $t$, yielding checkpoint answers $\{a^{(j)}_t\}_{j=1}^N$. These paths are clustered into groups $G = \{C_1, C_2, ..., C_k\}$ where $C_i = \{j | a^{(j)}_t = a_c\}$, forming answer-homogeneous clusters. 
The grouping results are marked with shading in Figure~\ref{fig:pipeline}. 

\textbf{3. Scoring.} A PRM assigns scores $s_j$ to each path $p_t^{(j)}$, with cluster $C_i$’s aggregate score computed as $S_i = \sum_{j \in C_i} s_j$. This approach is similar to a stepwise Weighted Best-of-N implementation.

\textbf{4. Selection.} Clusters are sorted by $S_i$ in descending order, while paths within each cluster are ranked by $s_j$. 
Then we sequentially select top-ranked paths across clusters via round-robin sampling: starting from the highest cluster, we pick the top path from each cluster, cycling back when reaching the last cluster until $M$ paths are selected.

This resource-aware branching prioritizes high-quality clusters while maintaining inter-cluster diversity. The cyclic selection mechanism prevents dominance by single-answer clusters and enables early identification of divergent reasoning trajectories. We provide a more rigorous process in Algorithm~\ref{alg:acs}.

\begin{algorithm}[t]
\SetAlgoLined
\small
\KwIn{Sampling budget $N$, beam width $M$, candidate set $\{p_t^{(j)}\}_{j=1}^N$}
\KwOut{Selected paths set $\mathcal{P}$}
\SetKwFunction{SRCA}{SRCA}
\SetKwFunction{PRM}{PRM}
% /* Checkpoint Injection \& Scoring */\;
\textcolor{mypink1}{\textit{// Checkpoint Injection \& Scoring}}

\For{$j \leftarrow 1$ \KwTo $N$}{
    $a^{(j)}_t \leftarrow$ \SRCA{$p_t^{(j)}$}\;
    $s^{(j)}_t \leftarrow$ \PRM{$p_t^{(j)}$}\;
}

\textcolor{mypink1}{\textit{// Clustering\& Sorting}}

$G = \{C_1, ..., C_k\}$ where $C_i = \{j | a^{(j)}_t = a_i\}$\;
\For{$C_i \in G$}{
    $S_i \leftarrow \sum_{j \in C_i} s_j$\;
}
% Sort clusters by $S_c$ in descending order\;
% Sort $\{C_i\}_{i=1}^n \text{ by } S_c \downarrow$\;
% $\text{sort}(\{C_i\}_{i=1}^n, S_c, \text{descending})$\;
$G \leftarrow \{C_i : S_{i} \geq S_{i+1}\}_{i=1}^k$\;
% $\{C_i\}_{i=1}^n \leftarrow \text{Sort }\{C_i\}_{i=1}^n \text{ by } S_c \downarrow$\;
\textcolor{mypink1}{\textit{// Round-robin Selection}}

$\mathcal{P}\leftarrow \emptyset$\;
\While{$|\mathcal{P}|< M$}{
    \For{$C_i\in G$}{
        $j^* \leftarrow \text{argmax}_{j \in C_i} s_j$\;
        $\mathcal{P} \leftarrow \mathcal{P} \cup \{p^{(j^*)}\}$\;
        $C_i \leftarrow C_i \setminus \{p^{(j^*)}\}$\;
        \If{$|\mathcal{P}|= M$}{
            break\;
        }
    }
}
\caption{Answer-Clustered Search}
\label{alg:acs}
\end{algorithm}

\vspace{-0.2cm}
\subsection{Checkpoint Candidate Augmentation}
The proposed Checkpoint Candidate Augmentation (CCA) aims to maximize the use of reasoning resources and enhance the diversity of candidate answers by integrating the checkpoint answers from intermediate reasoning steps. Traditional Beam Search methods retain only a fixed number, i.e., M, of complete reasoning paths as the final candidate set, which leads to the discard of many unfinished intermediate branches. To address this issue, our method continuously collects intermediate answers generated at all checkpoints during the ACS and reconstructs the corresponding truncated reasoning paths into valid candidate paths. Specifically, for each intermediate answer $a_t^{(j)}$ produced at a checkpoint, we concatenate it with the current reasoning path $p_t^{(j)}$ to form a candidate path with a complete logical chain:
\begin{equation}
    \hat{p}_t^{(j)} = p_t^{(j)} \oplus x_{\text{ckpt}} \oplus a_t^{(j)}
\end{equation}
where $\oplus$ represents string concatenation.

All candidate paths, including the original complete paths and the newly added intermediate paths, are uniformly scored by the PRM, and the path with the highest score is selected as the model output. This method offers two main advantages: first, by incorporating prediction results from the intermediate inference process into the candidate set, it significantly improves the utilization of computational resources already spent; second, by retaining intermediate answers at various stages, it establishes an effective fault tolerance mechanism. Even if the LLM makes mistakes in subsequent steps, it may still arrive at the correct answer through the retained high-quality intermediate predictions. 
On the other hand, CCA can effectively mitigate issues such as overthinking, increasingly erroneous reasoning, and repetitive outputs in LLMs.

% \usepackage[normalem]{ulem}
\begin{table*}[]
\centering
\small
\begin{tabular}{lcccc}
\hline
\multicolumn{1}{c}{\textbf{Models   \& TTS}} & \textbf{GSM8K} & \textbf{MATH500}     & \textbf{AIME}        & \textbf{OlympiadBench} \\ \hline
\rowcolor[HTML]{EFEFEF}\multicolumn{5}{c}{\textit{Independent Sampling}}                                                                                                    \\
Llama-3.1-70B-Instruct~\cite{grattafiori2024llama3herdmodels}                       & \textbf{95.10} & 65.00                & 36.66                & 27.70                  \\
Llama-3.2-1B-Instruct~\cite{llama32}                        & 43.75          & 24.40                & 3.22                 & 4.59                   \\
\hspace{1cm}\textit{w.} Self-Consistency~\cite{wang2023selfconsistency}                               & 57.70          & 39.80                & 8.57                 & 11.70                  \\
\rowcolor[HTML]{EFEFEF}\multicolumn{5}{c}{\textit{TTS Llama-3.2-1B-Instruct w. Llama3.1-8B-PRM-Deepseek-Data}}                                                                             \\
% Self-Consistency~\cite{wang2023selfconsistency}                             & 57.70          & 39.80                & 8.57                 & 11.70                  \\
BoN~\cite{brown2024largelanguagemonkeysscaling}                                          & 80.36          & 46.20                & 11.04                & 13.48                  \\
Weighted BoN~\cite{brown2024largelanguagemonkeysscaling}                                 & 65.50          & 46.40                & 10.50                & 13.63                  \\
Beam Search~\cite{snell2024scalingllmtesttimecompute}                                  & 84.84          & 52.00                & 19.07                & 18.07                  \\
DVTS~\cite{beeching2024scalingtesttimecompute}                                         & 83.47          & 52.60                & 20.68                & 19.40                  \\ \hdashline
SRCA (Ours)                                         & {\ul 85.60}    & {\ul 53.40}          & {\ul 24.97}          & {\ul 20.74}            \\
\rowcolor[HTML]{EFEFEF}\multicolumn{5}{c}{\textit{TTS Llama-3.2-1B-Instruct w. Skywork-o1-Open-PRM-Qwen-2.5-7B}}                                                                           \\
BoN~\cite{brown2024largelanguagemonkeysscaling}                                          & 80.74          & 55.20                & 25.08                & 18.22                  \\
Weighted BoN~\cite{brown2024largelanguagemonkeysscaling}                                 & 76.72          & 52.60                & 28.08                & 18.67                  \\
Beam Search~\cite{snell2024scalingllmtesttimecompute}                                  & 84.99          & 63.20                & 26.82                & 23.89                  \\
DVTS~\cite{beeching2024scalingtesttimecompute}                                         & 84.00          & 64.80                & 29.03                & 25.82                  \\ \hdashline
SRCA (Ours)                                         & {\ul 85.97}    & {\ul \textbf{65.20}} & {\ul \textbf{39.71}} & {\ul \textbf{27.75}}   \\ \hline
\end{tabular}
\caption{Comparison of TTS results. 
% The upper section presents results from larger models under greedy search settings. 
% The upper section reports results from 1B and 72B models under greedy search for comparison. Below are results from the 1B model combined with various TTS methods and two PRMs, with sampling size N=128.
The upper section shows the greedy search results for 1B and 70B models, and we additionally report the self-consistency performance of the 1B model with $N=128$.
The lower section shows results from the 1B model combined with various TTS methods and two PRMs, also with $N=128$. 
Numbers indicate accuracy (\%).
Best overall performance on each dataset is marked in \textbf{bold}, while best performance within each group is {\ul underlined}.}
\label{tab:tab1_main_res}
% \vspace{-0.3cm}
\end{table*}
\section{Experiments}
We conducted comparative experiments on four mathematical reasoning datasets and against four Test-Time Scaling baselines. 
\subsection{Settings}
In the experiments, we used four datasets: GSM8K, MATH500, AIME, and OlympiadBench.
Two different-sized LLMs were tested in total, specifically \texttt{Llama}-\texttt{3.2}-\texttt{1B}-\texttt{Instruct}~\cite{llama32} and \texttt{Qwen3}-\texttt{0.6B}~\cite{yang2025qwen3technicalreport}. 
For the PRM, we adopted the model fine-tuned by DeepSeek, \texttt{Llama3.1}-\texttt{8B}-\texttt{PRM}-\texttt{Deepseek}-\texttt{Data}~\cite{xiong2024rlhflowmath} and \texttt{Skywork}-\texttt{o1}-\texttt{Open}-\texttt{PRM}-\texttt{Qwen-2.5}-\texttt{7B} released by Skywork~\cite{skyworkopeno12024}. 

We compared our method against several TTS algorithms, including Greedy Search, Self-Consistency~\cite{wang2023selfconsistency}, Best-of-N (BoN)~\cite{brown2024largelanguagemonkeysscaling}, Weighted BoN~\cite{brown2024largelanguagemonkeysscaling}, Beam Search~\cite{snell2024scalingllmtesttimecompute}, and Diverse Verifier Tree Search (DVTS)~\cite{beeching2024scalingtesttimecompute}. Among these methods, Beam Search maintains $N$ paths and selects $M$ highest-scoring ones for expansion, with each generating $N/M$ sub-paths. DVTS extends this by initializing $M$ subtrees and sampling $N/M$ paths per step within each subtree, enhancing path diversity through subtree isolation. Other baseline methods are standard approaches in the field; their detailed descriptions can be found in Appendix~\ref{app_sec:exp_settings}.

For all the sampling-based methods, set temperature = 0.8 and top\_p = 0.9. We use $N=16$ and $N=64$ for sampling times to assess the effect of sampling scale on reasoning performance. For the methods involving path selection, such as Beam Search, DVTS, and SRCA, the beam width $M$ is fixedly set to 4, that is, the 4 candidate paths with the highest scores are retained at each reasoning step for subsequent expansion.

Since the PRM can collect the step-level scores of the complete reasoning path to form a score sequence, there are various ways to determine the final score of the path, such as taking the sum, accumulation, minimum value of the sequence scores, and the score of the last step. In this experiment, the score of the last step in the path is used as the path score. The effects of these configurations on experimental results are discussed in Appendix~\ref{app_sec:reduction_bon}.

\subsection{Results}
\subsubsection{Scaling with SRCA: Small Models Can Outperform Larger Ones}

Table~\ref{tab:tab1_main_res} shows the performance of \texttt{Llama}-\texttt{3.2}-\texttt{1B}-\texttt{Instruct} with various TTS methods on four mathematical datasets ($N=128$, $M=4$). We also include results from \texttt{Llama}-\texttt{3.1}-\texttt{70B}-\texttt{Instruct} for comparison.

SRCA consistently outperforms other TTS methods across all datasets, regardless of the PRM used. With DeepSeek PRM, SRCA achieves approximately 10\% absolute improvement over the BoN baseline. The improvement is particularly notable on AIME, where SRCA shows a 43\% relative performance gain over DVTS.
Remarkably, when using Skywork PRM, our 1B model with SRCA outperforms the 70B model on MATH500, AIME, and OlympiadBench, only falling behind on the simpler GSM8K dataset. This demonstrates SRCA's effectiveness in enabling smaller models to compete with larger ones.
On the other hand, The choice of PRM also impacts performance, with Skywork PRM generally yielding better results than DeepSeek PRM across all TTS methods. This suggests that future advances in PRM development could lead to further performance improvements in TTS methods. More results of SRCA on Qwen3-0.6B are presented in Appendix~\ref{app_sec:qwen3}.

\subsubsection{Expanding Sampling Times: SRCA Has Higher Efficiency.}

\begin{figure*}[ht]
    \centering
    \includegraphics[width=0.9\linewidth]{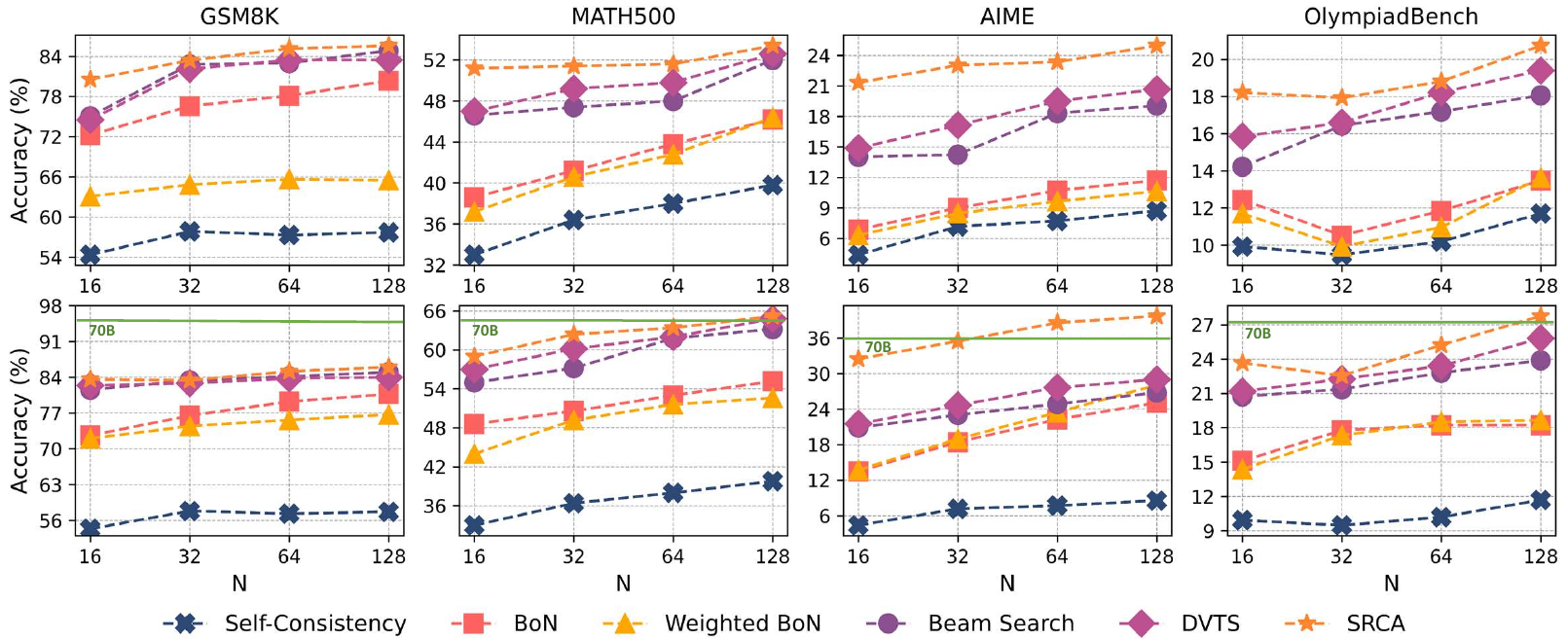}
    \caption{Performance trends of TTS methods with DeepSeek PRM (top row) and Skywork PRM (bottom row) and  as the sampling number $N$ increases from 16 to 128. In the bottom row, we additionally mark the performance of the 70B model with a green line for comparison.}
    \label{fig:N_plots}
\end{figure*}

\begin{figure*}[ht]
    \centering
    \includegraphics[width=0.9\linewidth]{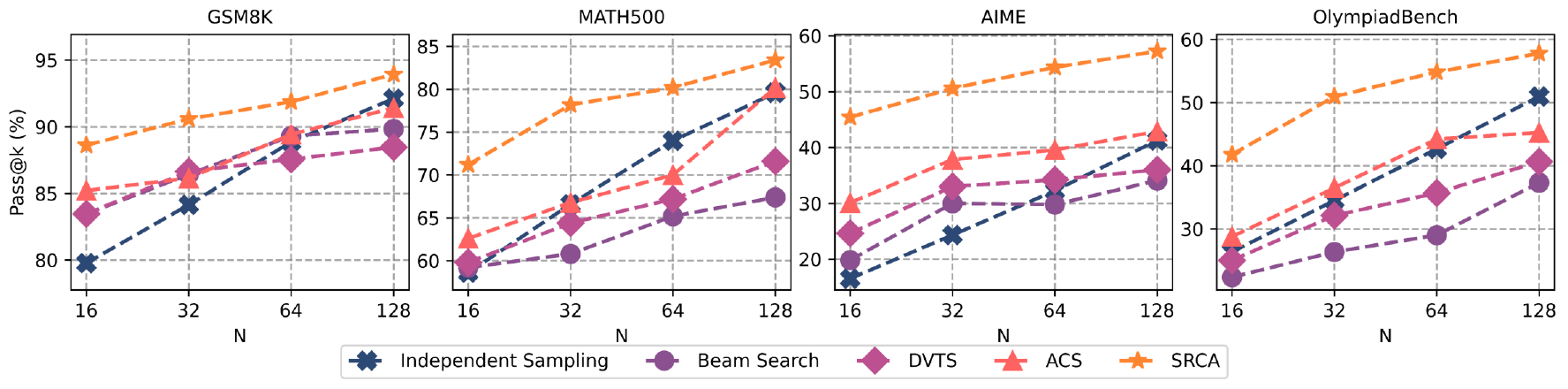}
    \caption{Pass@K trends of the 1B model with different TTS methods and DeepSeek PRM as the sampling number increases from 16 to 128. Note that for Pass@K calculation, Self-Consistency, BoN, and Weighted BoN degrade to Independent Sampling.}
    \label{fig:pass_k}
    \vspace{-0.4cm}
\end{figure*}

\noindent We test various TTS methods with sampling times $N = 16$, $32$, $64$, and $128$. The results are shown in Figure~\ref{fig:N_plots}.
SRCA demonstrates superior efficiency by requiring fewer samples to achieve comparable accuracy. With DeepSeek PRM on MATH500, SRCA achieves 51.2\% accuracy at $N=16$, outperforming DVTS's 49.8\% at $N=64$. This advantage is more pronounced on AIME, where SRCA's accuracy at $N=16$ exceeds all TTS methods' performance at $N=128$. Using Skywork PRM further amplifies this gap: SRCA reaches 32.48\% at $N=16$, while the best baseline (DVTS) only achieves 29.03\% at $N=128$.
Performance improvements show diminishing returns as $N$ increases, with $N = 16\rightarrow32$ gains being larger than $N = 64\rightarrow128$. This pattern holds across different PRMs, suggesting convergence to an upper bound. 
Further analysis regarding the computational overhead and efficiency of SRCA is provided in Appendix~\ref{app_sec:flops}.

\vspace{-0.3cm}
\section{Analysis}
\vspace{-0.2cm}
\subsection{Pass Rate Test: SRCA Improves Answer Discovery}

The ability to sample at least one correct reasoning path is crucial for policy models, as it determines the effectiveness of PRM guidance. If a policy model does not sample any correct paths, even a perfect PRM cannot select the correct one. We conducted \textbf{Pass@k} tests on 4 datasets comparing different TTS methods, including SRCA without CCA to understand each component's contribution. Results are shown in Figure~\ref{fig:pass_k}.
SRCA demonstrates superior pass rates across datasets. Ablation studies show that CCA contributes approximately 10\% improvement through answer pool expansion, while ACS alone still outperforms DVTS and Beam Search by 3\% on average.
Independent sampling achieves higher Pass@k on simpler datasets due to its unconstrained randomness generating more diverse solutions. However, for challenging datasets like AIME, this approach performs poorly as random exploration is ineffective when precise reasoning is required. On the other hand, the TTS method produces a candidate set of better quality, resulting in a higher pass rate.

\begin{figure*}[ht]
    \centering
    \includegraphics[width=0.9\linewidth]{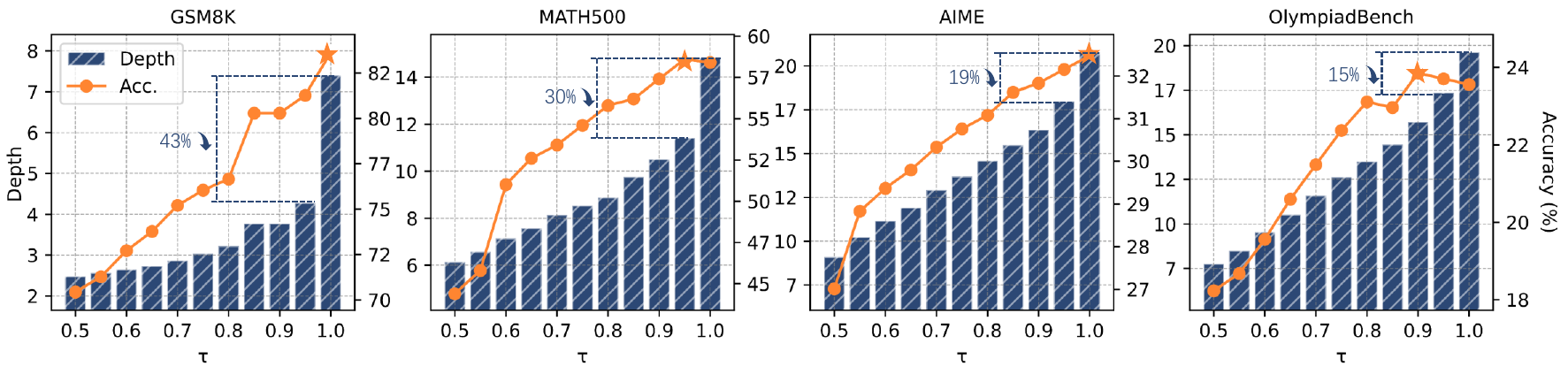}
    \caption{The average accuracy and search depth of SRCA with early stopping strategies under different values of tau. The left y-axis represents the search depth, while the right y-axis represents the accuracy (\%). The dashed line in the figure annotates the reduction rate of tree depth, i.e., the number of reasoning steps, when $tau=0.95$. The pentagon represents the best performance.}
    \label{fig:early_stop}
\end{figure*}

\begin{figure*}[ht]
    \centering
    \includegraphics[width=0.9\linewidth]{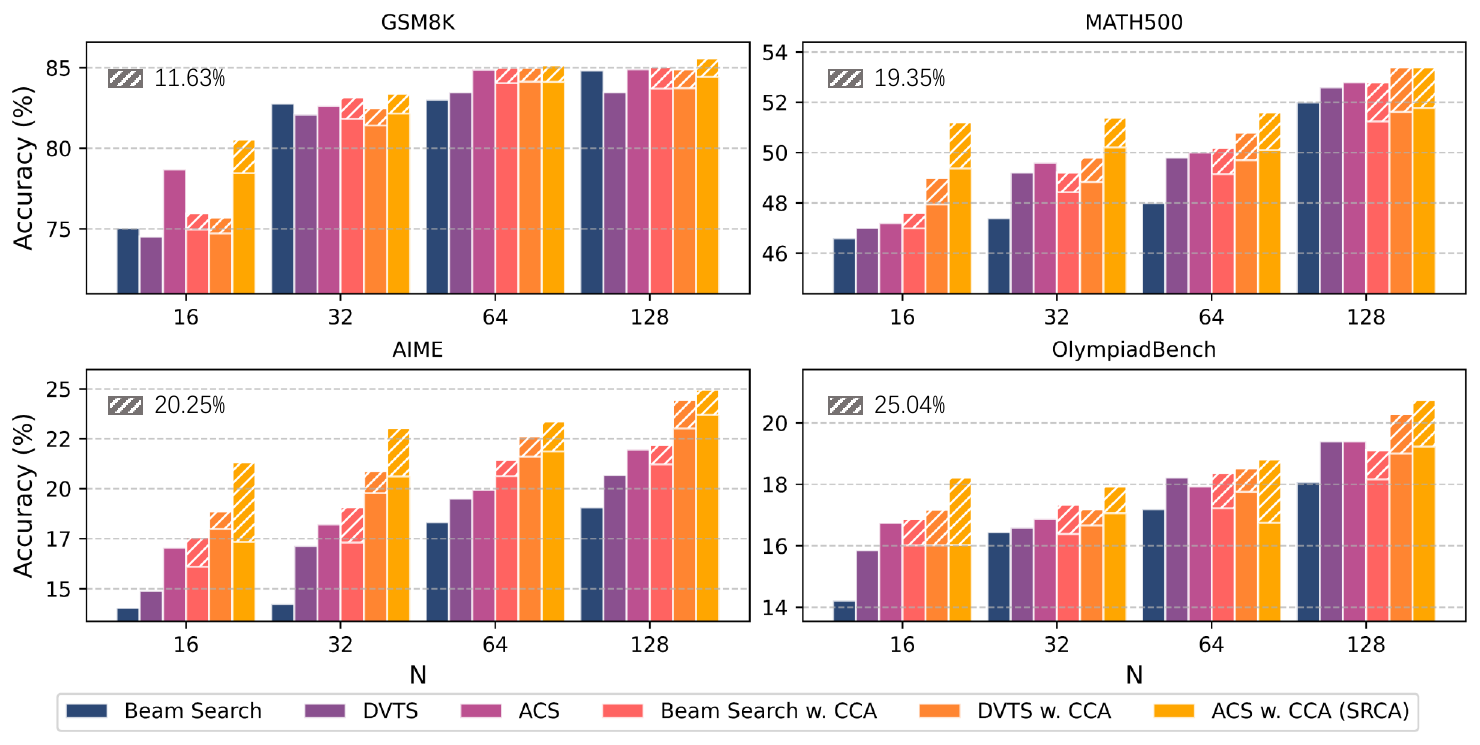}
    \caption{Ablation study results on four datasets, grouped by different values of $N$. For the bars corresponding to methods incorporating CCA, the Checkpoint Answer Rate (CAR) is additionally marked with slashes shading. The average CAR for each dataset is indicated in the top-left corner of each subplot.}
    \label{fig:ablation}
    \vspace{-0.3cm}
\end{figure*}

\subsection{Early Stopping: Efficient Computing with No Performance Loss.}
\vspace{-0.2cm}
Recent research shows that LLMs often suffer from overthinking, conducting unnecessary analysis that wastes computational resources and can even lead to incorrect answers~\cite{li2024escapeskyhighcostearlystopping, wu2025lessunderstandingchainofthoughtlength, sui2025stopoverthinkingsurveyefficient}.
We implement \textbf{early stopping} in SRCA by introducing a threshold $\tau$: reasoning stops when a checkpoint answer's score exceeds $\tau$. We tested various $\tau$ values (0.5-1.0), measuring both accuracy and average reasoning steps, with $\tau=1$ (no early stopping) as baseline. Results are shown in Figure~\ref{fig:early_stop}. This experiment uses Skywork PRM with $N=16$ samples.

Low thresholds like $0.5$ hurt performance, causing a 14\% accuracy drop on MATH500. While this reduces reasoning steps, the performance trade-off is unacceptable. Higher thresholds require more steps but yield better accuracy, as expected.
At $\tau=0.95$, early stopping reduces reasoning steps by 27\% across datasets while only losing 0.58\% accuracy. Notably, it sometimes improves accuracy: both MATH500 and OlympiadBench show better results at $\tau=0.95$ than without early stopping. This confirms that overthinking can harm performance, due to either model reasoning limitations or PRM imperfections~\cite{zheng2024processbenchidentifyingprocesserrors, he2025largelanguagemodelsdetect}. Early stopping is particularly effective for simpler tasks: GSM8K shows a 43\% reduction in reasoning steps at $\tau=0.95$, compared to 19\% for the more challenging AIME problems.

\subsection{Ablation Study}
\begin{table*}[ht]
    \centering
    \includegraphics[scale=1.0, width=1.0\textwidth]{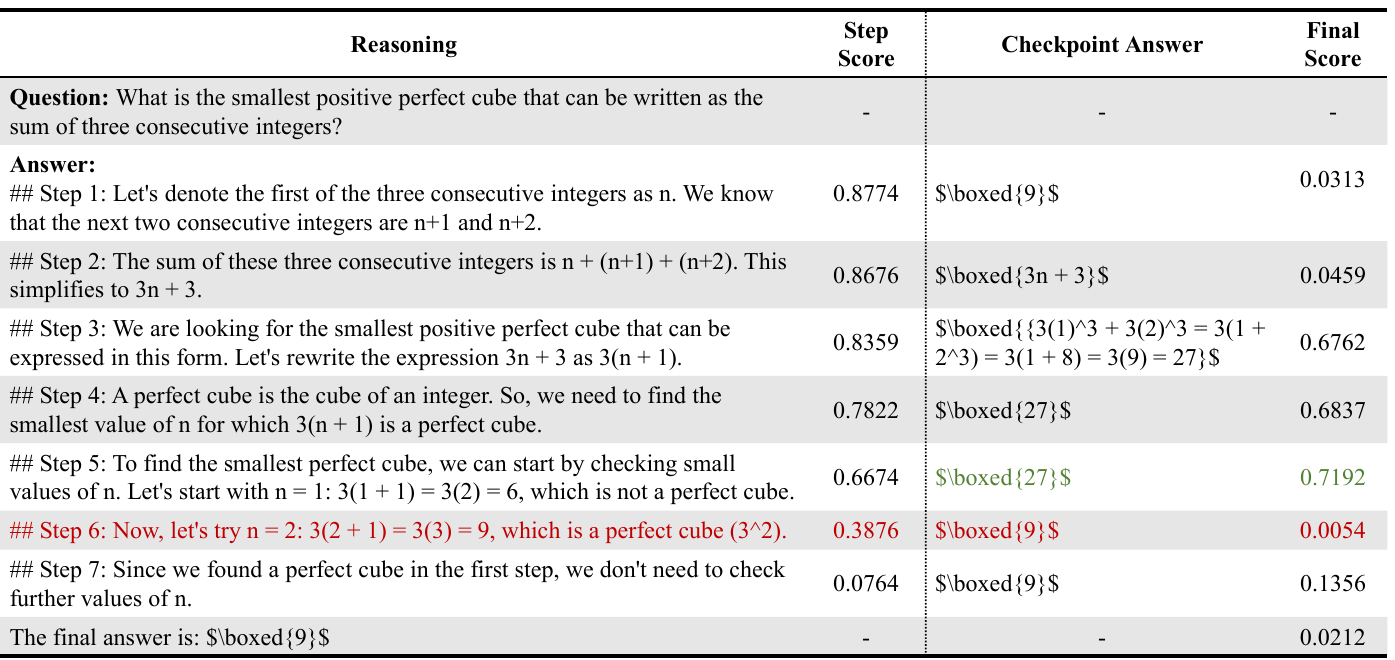}
    \caption{A real running case of SRCA during reasoning. The first column shows the question and the model's reasoning process.
    % For better clarity and readability, we've omitted the conversation template from the data. 
     The Step Score represents PRM's evaluation of the current step, while the Final Score indicates the PRM score when using the current step and its checkpoint answer as the reasoning endpoint.
    }
    \label{tab:case_study}
    \vspace{-0.2cm}
\end{table*}

SRCA combines ACS for diverse path searching and CCA for expanding the answer candidate pool. We conducted ablation studies by: (1) testing ACS alone without CCA, and (2) combining CCA with baseline methods (Beam Search and DVTS). Results are shown in Figure 1. We also track the Checkpoint Answer Rate (CAR) - the percentage of final answers selected from checkpoint, which is marked with slashes on the bar chart.

After removing the CCA strategy, SRCA degrades to ACS strategy, resulting in a notable performance decline. However, on relatively simple datasets like GSM8K and MATH500, increasing the sampling size (e.g., to $N=128$) minimizes this performance gap to less than 1\%. Notably, ACS consistently outperforms both Beam Search and DVTS baselines across most configurations, demonstrating its robust effectiveness.

The integration of CCA with baseline methods yields substantial improvements in accuracy (3-4\%). Analysis shows that 19.07\% of final answers originate from CCA's expanded candidate pool, underscoring its important contribution to solution generation. The impact varies by problem difficulty: CAR is 11.63\% for GSM8K but rises to 25.04\% for OlympiadBench, indicating that CCA's influence is more pronounced in solving complex problems.

\subsection{Case Study}
Table~\ref{tab:case_study} shows a real running case of SRCA during reasoning. Since the complete search tree is too large, we only showcase how SRCA uses Checkpoint Answers to backup correct answers from incorrect branches. The model's reasoning process can be explained manually in three phases:\vspace{-0.2cm}
\begin{itemize}
    \item Early Stage (Steps 1-2): During initial reasoning, the model produces either incorrect answers or incomplete expressions instead of proper numerical values, indicating insufficient reasoning depth. \vspace{-0.2cm}
    \item Answer Formation Stage (Steps 3-5): Starting from Step 3, the model attempts brief reasoning in the answer box and first obtains the correct answer 27. Although reasoning in the answer box is not ideal behavior, the model successfully reaches the correct answer this way. This correct answer is maintained until Step 5.\vspace{-0.2cm}
    \item Error Stage (Step 6): A critical reasoning error occurs when the model incorrectly identifies 9 as a perfect cube instead of a perfect square. This error leads to an incorrect checkpoint answer that affects the reasoning until the end.\vspace{-0.2cm}
\end{itemize}

When all checkpoint answers are evaluated as reasoning endpoints by PRM, Step 5 receives the highest score of 0.7192, exceeding the natural ending's score of 0.0212. If no other branch has a score higher than 0.7192, this score will be selected as the final answer, effectively correcting the wrong answer from the natural reasoning endpoint to the correct one.

\section{Conclusion}

In this paper, we introduced SRCA, a novel framework that enhances LLM reasoning by introducing checkpoints between reasoning steps. Our Answer-Clustered Search strategy effectively maintains path diversity while ensuring reasoning quality, and the Checkpoint Candidate Augmentation approach efficiently utilizes intermediate predictions for final decision-making. Experimental results demonstrate that SRCA outperforms baseline methods like Beam Search and DVTS across various mathematical reasoning datasets. The success of SRCA suggests that leveraging intermediate checkpoints is a promising direction for improving LLM reasoning capabilities.

\section{Limitations}
SRCA faces two main limitations. First, while it requires checkpoints between reasoning steps, defining precise step boundaries is challenging.
Although Llama-series models exhibit relatively clear step demarcations with characteristic delimiters, others, 
particularly the emerging "slow thinking" models, often generate outputs without distinct structural patterns and sometimes in a more conversational style. Second, the reasoning steps augmented by the CCA strategy are often incomplete. While models can still generate correct answers based on these partial reasoning paths, this incompleteness reduces the interpretability of the reasoning process. Compared to naturally completed reasoning chains, these occasionally truncated paths represent a shortcoming in terms of explanation quality and transparency.

\bibliography{acl_latex}
\clearpage
\appendix
\section{Experiment Settings}\label{app_sec:exp_settings}

\subsection{Datasets}
The following are the four datasets used in the experiment:\vspace{-0.2cm}
\begin{itemize}
    \item \textbf{GSM8K}~\cite{cobbe2021trainingverifierssolvemath} is an evaluation set consisting of 8,500 high-quality primary school mathematics problems. It is mainly used to assess the language comprehension and mathematical reasoning abilities of models in basic mathematical problems.\vspace{-0.2cm}
    \item \textbf{MATH500}~\cite{lightman2024lets} is a subset of the MATH dataset~\cite{hendrycks2021measuring} containing 500 questions. It covers seven mathematical domains and five difficulty levels. It is designed to test the performance of LLMs in solving advanced mathematical problems.\vspace{-0.2cm}
    \item \textbf{AIME}\footnote{{www.kaggle.com/datasets/aime-problem-set-1983-2024}} offers a rich collection of challenging problems from the American Invitational Mathematics Examination and contains 933 high-difficulty mathematical problems.\vspace{-0.2cm}
    \item \textbf{OlympiadBench} is an Olympiad-level bilingual multimodal scientific benchmark~\cite{he-etal-2024-olympiadbench}. In this experiment, only the subset of English mathematical problems without images was tested, with a total of 674 questions.  \vspace{-0.2cm}
\end{itemize}

\subsection{Baselines}
\begin{itemize}
    \item Greedy Search: A decoding method based on the principle of local optimality. It always selects the token with the highest current probability as the output.\vspace{-0.3cm}
    \item Self-Consistency~\cite{wang2023selfconsistency}: LLMs generates $N$ independent reasoning paths. Finally, the most frequently occurring output is counted as the answer.\vspace{-0.3cm}
    \item Best-of-N (BoN)~\cite{brown2024largelanguagemonkeysscaling}: Similar to self-consistency, the LLM generates $N$ independent reasoning paths. According to the scores given by the reward model, the path with the highest score is selected as the answer.\vspace{-0.3cm}
    \item Weighted BoN~\cite{brown2024largelanguagemonkeysscaling}: It is a combination of Self-Consistency and BoN. The reward model scores the $N$ independent reasoning paths generated by the LLM. Then, the paths are clustered according to the answers, and the sum of the path scores within each cluster is taken as the answer's score. The answer with the highest score is selected.\vspace{-0.3cm}
    \item Beam Search~\cite{snell2024scalingllmtesttimecompute}: $N$ reasoning paths are maintained at each reasoning step. According to the scores given by the PRM for the current paths, $M$ paths are selected to continue the reasoning and expand downward. Each selected path can expand into $N/M$ sub-paths. \vspace{-0.3cm}
    \item Diverse Verifier Tree Search (DVTS)~\cite{beeching2024scalingtesttimecompute}: DVTS is an extension of the beam search. It first initializes $M$ subtrees. Each subtree samples $N/M$ paths at every step. Those paths are then scored by the PRM. The path with the highest score within the subtree is selected for further reasoning. It is similar to Beam Search, as in each step, M paths are selected from N paths for further reasoning. Due to the scope limitation of the subtrees, it prevents some locally optimal branches from early elimination, thereby enhancing the path diversity.   \vspace{-0.3cm}
\end{itemize}

\section{Supplementary Experimental Results}
\subsection{Computational Cost Analysis in FLOPs}
\label{app_sec:flops}
\begin{table*}[ht]
\centering
\begin{tabular}{llll}
\hline
\multicolumn{1}{c}{\textbf{Model}} & \multicolumn{1}{c}{\textbf{Type}} & \multicolumn{1}{c}{\textbf{N}} & \multicolumn{1}{c}{\textbf{FLOPs}} \\ \hline
Llama-3.2-1B-Instruct              & Auto Regressive                   & 128                            & $1.31 \times 10^{18}$                           \\
Llama3.1-8B-PRM-Deepseek-Data      & Reward                            & 128                            & $9.03 \times 10^{15}$                           \\
Llama-3.1-70B-Instruc              & Auto Regressive                   & 1                              & $3.04 \times 10^{18}$                           \\ \hline
\end{tabular}
\caption{Computational Cost Analysis (in FLOPs) for Different Model Configurations during Inference.}
\label{tab:flops}
\end{table*}
Table~\ref{tab:flops} compares the computational cost in FLOPs for processing a single query across different models. We assume an input length of 256 tokens (prefill) and an output length of 4096 tokens (decode). The policy model generates tokens sequentially in an auto-regressive manner, requiring multiple forward passes, while the PRM requires only one forward pass for scoring.

The results demonstrate that SRCA with $N=128$, combining a 1B policy model and an 8B PRM, requires only 43.01\% of the computational cost compared to the 70B model for processing a single sample. Considering the experimental results reported in Table~\ref{tab:tab1_main_res}, the 1B model enhanced with SRCA achieves higher accuracy than the 70B model, indicating that our approach not only reduces computational overhead but also yields superior performance.

\subsection{Evaluation of Scoring Methods and Selection Strategies}
\label{app_sec:reduction_bon}

% \usepackage{multirow}
\begin{table*}[ht]
\centering
\setlength\tabcolsep{2pt}
\renewcommand{\arraystretch}{1.2}

\small
\begin{tabular}{cclcccccccccc}
\hline
\multirow{2}{*}{\textbf{Selection}}                                                                & \multirow{2}{*}{\textbf{N}} & \multicolumn{1}{c}{\multirow{2}{*}{\textbf{Method}}} & \textbf{GSM8k} & \textbf{MATH500} & \textbf{AIME} & \textbf{Olympiad} & \multirow{2}{*}{\textbf{Avg.}} & \textbf{GSM8k} & \textbf{MATH500} & \textbf{AIME} & \textbf{Olympiad} & \multirow{2}{*}{\textbf{Avg.}} \\ \cmidrule(lr){4-7} \cmidrule(lr){9-12}
                                                                                                   &                             & \multicolumn{1}{c}{}                                 & \multicolumn{4}{c}{Last}                                                   &                                & \multicolumn{4}{c}{Mean}                                                   &                                \\ \hline
\multicolumn{1}{c|}{\multirow{16}{*}{BoN}}                                                         & \multirow{4}{*}{16}         & Beam                                          & 0.7505         & 0.4660           & 0.1404        & 0.1422                 & \multicolumn{1}{c|}{0.3748}    & 0.7475         & 0.4600           & 0.1200        & 0.1247                 & 0.3631                         \\
\multicolumn{1}{c|}{}                                                                              &                             & DVTS                                                 & 0.7452         & 0.4700           & 0.1489        & 0.1585                 & \multicolumn{1}{c|}{0.3807}    & 0.7331         & 0.4620           & 0.1307        & 0.1525                 & 0.3696                         \\
\multicolumn{1}{c|}{}                                                                              &                             & SRCA                                                 & 0.8054         & 0.5120           & 0.2133        & 0.1822                 & \multicolumn{1}{c|}{0.4282}    & 0.7869         & 0.4940           & 0.1747        & 0.1718                 & 0.4069                         \\
\multicolumn{1}{c|}{}                                                                              &                             & Avg.                                                 & 0.7670         & 0.4827           & 0.1675        & 0.1610                 & \multicolumn{1}{c|}{\cellcolor[HTML]{FFCCCC}{0.3946}}    & 0.7558         & 0.4720           & 0.1418        & 0.1497                 & \cellcolor[HTML]{FFCCCC}0.3798                         \\ \cdashline{2-13}
\multicolumn{1}{c|}{}                                                                              & \multirow{4}{*}{32}         & Beam                                          & 0.8278         & 0.4740           & 0.1425        & 0.1644                 & \multicolumn{1}{c|}{0.4022}    & 0.8043         & 0.4780           & 0.1457        & 0.1377                 & 0.3914                         \\
\multicolumn{1}{c|}{}                                                                              &                             & DVTS                                                 & 0.8210         & 0.4920           & 0.1714        & 0.1659                 & \multicolumn{1}{c|}{0.4126}    & 0.8241         & 0.5020           & 0.1758        & 0.1629                 & 0.4162                         \\
\multicolumn{1}{c|}{}                                                                              &                             & SRCA                                                 & 0.8340         & 0.5140           & 0.2304        & 0.1793                 & \multicolumn{1}{c|}{0.4394}    & 0.8317         & 0.5120           & 0.1908        & 0.1793                 & 0.4285                         \\
\multicolumn{1}{c|}{}                                                                              &                             & Avg.                                                 & 0.8276         & 0.4933           & 0.1814        & 0.1699                 & \multicolumn{1}{c|}{\cellcolor[HTML]{FFCCCC}0.4181}    & 0.8200         & 0.4973           & 0.1708        & 0.1600                 & \cellcolor[HTML]{FFCCCC}0.4120                         \\ \cdashline{2-13}
\multicolumn{1}{c|}{}                                                                              & \multirow{4}{*}{64}         & Beam                                          & 0.8302         & 0.4800           & 0.1833        & 0.1719                 & \multicolumn{1}{c|}{0.4164}    & 0.8392         & 0.4960           & 0.1758        & 0.1659                 & 0.4192                         \\
\multicolumn{1}{c|}{}                                                                              &                             & DVTS                                                 & 0.8347         & 0.4980           & 0.1951        & 0.1822                 & \multicolumn{1}{c|}{0.4275}    & 0.8484         & 0.5140           & 0.1907        & 0.1733                 & 0.4316                         \\
\multicolumn{1}{c|}{}                                                                              &                             & SRCA                                                 & 0.8514         & 0.5160           & 0.2337        & 0.1881                 & \multicolumn{1}{c|}{0.4473}    & 0.8491         & 0.5160           & 0.2144        & 0.1837                 & 0.4408                         \\
\multicolumn{1}{c|}{}                                                                              &                             & Avg.                                                 & 0.8388         & 0.4980           & 0.2040        & 0.1807                 & \multicolumn{1}{c|}{\cellcolor[HTML]{FFCCCC}0.4304}    & 0.8456         & 0.5087           & 0.1936        & 0.1743                 & \cellcolor[HTML]{FFCCCC}0.4305                         \\ \cdashline{2-13}
\multicolumn{1}{c|}{}                                                                              & \multirow{4}{*}{128}        & Beam                                          & 0.8484         & 0.5200           & 0.1907        & 0.1807                 & \multicolumn{1}{c|}{0.4350}    & 0.8340         & 0.5160           & 0.1832        & 0.1807                 & 0.4285                         \\
\multicolumn{1}{c|}{}                                                                              &                             & DVTS                                                 & 0.8347         & 0.5260           & 0.2068        & 0.1940                 & \multicolumn{1}{c|}{0.4404}    & 0.8499         & 0.5180           & 0.1843        & 0.1866                 & 0.4347                         \\
\multicolumn{1}{c|}{}                                                                              &                             & SRCA                                                 & 0.8560         & 0.5340           & 0.2497        & 0.2074                 & \multicolumn{1}{c|}{0.4618}    & 0.8514         & 0.5240           & 0.2197        & 0.1896                 & 0.4462                         \\
\multicolumn{1}{c|}{}                                                                              &                             & Avg.                                                 & 0.8464         & 0.5267           & 0.2157        & 0.1940                 & \multicolumn{1}{c|}{\cellcolor[HTML]{FFCCCC}0.4457}    & 0.8451         & 0.5193           & 0.1957        & 0.1856                 & \cellcolor[HTML]{FFCCCC}0.4365                         \\ \hline
\multicolumn{1}{c|}{\multirow{16}{*}{\begin{tabular}[c]{@{}c@{}}Weighted\\      BoN\end{tabular}}} & \multirow{4}{*}{16}         & Beam                                          & 0.7369         & 0.4600           & 0.1200        & 0.1303                 & \multicolumn{1}{c|}{0.3618}    & 0.7194         & 0.4460           & 0.1189        & 0.1229                 & 0.3518                         \\
\multicolumn{1}{c|}{}                                                                              &                             & DVTS                                                 & 0.7422         & 0.4760           & 0.1446        & 0.1526                 & \multicolumn{1}{c|}{0.3789}    & 0.7111         & 0.4620           & 0.1125        & 0.1496                 & 0.3588                         \\
\multicolumn{1}{c|}{}                                                                              &                             & SRCA                                                 & 0.7597         & 0.4800           & 0.1714        & 0.1688                 & \multicolumn{1}{c|}{0.3950}    & 0.7187         & 0.4680           & 0.1393        & 0.1674                 & 0.3734                         \\
\multicolumn{1}{c|}{}                                                                              &                             & Avg.                                                 & 0.7463         & 0.4720           & 0.1453        & 0.1506                 & \multicolumn{1}{c|}{\cellcolor[HTML]{FFCCCC}0.3785}    & 0.7164         & 0.4587           & 0.1236        & 0.1466                 & \cellcolor[HTML]{FFCCCC}0.3613                         \\ \cdashline{2-13}
\multicolumn{1}{c|}{}                                                                              & \multirow{4}{*}{32}         & Beam                                          & 0.7740         & 0.4760           & 0.1446        & 0.1659                 & \multicolumn{1}{c|}{0.3901}    & 0.7520         & 0.4480           & 0.1404        & 0.1348                 & 0.3688                         \\
\multicolumn{1}{c|}{}                                                                              &                             & DVTS                                                 & 0.7877         & 0.4780           & 0.1661        & 0.1718                 & \multicolumn{1}{c|}{0.4009}    & 0.7491         & 0.4680           & 0.1425        & 0.1644                 & 0.3810                         \\
\multicolumn{1}{c|}{}                                                                              &                             & SRCA                                                 & 0.7937         & 0.4900           & 0.1822        & 0.1778                 & \multicolumn{1}{c|}{0.4109}    & 0.7832         & 0.4700           & 0.1704        & 0.1762                 & 0.4000                         \\
\multicolumn{1}{c|}{}                                                                              &                             & Avg.                                                 & 0.7851         & 0.4813           & 0.1643        & 0.1718                 & \multicolumn{1}{c|}{\cellcolor[HTML]{FFCCCC}0.4007}    & 0.7614         & 0.4620           & 0.1511        & 0.1585                 & \cellcolor[HTML]{FFCCCC}0.3833                         \\ \cdashline{2-13}
\multicolumn{1}{c|}{}                                                                              & \multirow{4}{*}{64}         & Beam                                          & 0.8036         & 0.4780           & 0.1886        & 0.1733                 & \multicolumn{1}{c|}{0.4109}    & 0.7771         & 0.4860           & 0.1939        & 0.1615                 & 0.4046                         \\
\multicolumn{1}{c|}{}                                                                              &                             & DVTS                                                 & 0.7915         & 0.4860           & 0.1897        & 0.1825                 & \multicolumn{1}{c|}{0.4124}    & 0.7839         & 0.4900           & 0.1961        & 0.1719                 & 0.4105                         \\
\multicolumn{1}{c|}{}                                                                              &                             & SRCA                                                 & 0.8173         & 0.5060           & 0.2068        & 0.1854                 & \multicolumn{1}{c|}{0.4289}    & 0.7945         & 0.4940           & 0.2208        & 0.1778                 & 0.4218                         \\
\multicolumn{1}{c|}{}                                                                              &                             & Avg.                                                 & 0.8041         & 0.4900           & 0.1950        & 0.1804                 & \multicolumn{1}{c|}{\cellcolor[HTML]{FFCCCC}0.4174}    & 0.7852         & 0.4900           & 0.2036        & 0.1704                 & \cellcolor[HTML]{FFCCCC}0.4123                         \\ \cdashline{2-13}
\multicolumn{1}{c|}{}                                                                              & \multirow{4}{*}{128}        & Beam                                          & 0.8014         & 0.5000           & 0.1907        & 0.1854                 & \multicolumn{1}{c|}{0.4194}    & 0.7574         & 0.4820           & 0.1951        & 0.1911                 & 0.4064                         \\
\multicolumn{1}{c|}{}                                                                              &                             & DVTS                                                 & 0.8195         & 0.5020           & 0.2079        & 0.1899                 & \multicolumn{1}{c|}{0.4298}    & 0.7680         & 0.4920           & 0.1994        & 0.1940                 & 0.4134                         \\
\multicolumn{1}{c|}{}                                                                              &                             & SRCA                                                 & 0.8173         & 0.5100           & 0.2262        & 0.1943                 & \multicolumn{1}{c|}{0.4370}    & 0.7786         & 0.5120           & 0.2444        & 0.2030                 & 0.4345                         \\
\multicolumn{1}{c|}{}                                                                              &                             & Avg.                                                 & 0.8127         & 0.5040           & 0.2083        & 0.1899                 & \multicolumn{1}{c|}{\cellcolor[HTML]{FFCCCC}0.4287}    & 0.7680         & 0.4953           & 0.2130        & 0.1960                 & \cellcolor[HTML]{FFCCCC}0.4181                         \\ \hline
\end{tabular}
\caption{Performance comparison of TTS methods with different scoring methods (Last/Mean) and selection strategies (BoN/Weighted BoN) on four benchmark datasets. Numbers indicate accuracy. Higher scores indicate better performance. Red cells denote group averages for each N value.}
\label{tab:reduction_and_bon}
\end{table*}

We analyzed how different scoring methods for reasoning paths and answer selection strategies affect the accuracy of TTS methods. 
We employ \texttt{Llama3.1}-\texttt{8B}-\texttt{PRM}-\texttt{Deepseek}-\texttt{Data} as the PRM in this experiment.
The PRM assigns scores to each reasoning step, generating a sequence of scores for each path. We examined two primary methods for computing the final path score:
\begin{itemize}\vspace{-0.2cm}
    \item \textbf{Last}: Using the final step's score as the path score, where PRM functions similarly to an Outcome Reward Model (ORM). \vspace{-0.2cm}
    \item \textbf{Mean}: Taking the average of the score sequence to reflect the overall reliability of the reasoning process.
\end{itemize}

Furthermore, BoN and Weighted BoN can be combined with other tree search algorithms as answer selection strategies. Specifically, after the tree search algorithm generates multiple candidate paths:
\begin{itemize}\vspace{-0.2cm}
\item \textbf{BoN}: Selects the path with the highest score\vspace{-0.2cm}
\item \textbf{Weighted BoN}: First clusters answers, then selects the answer with the highest sum of path scores within its cluster
\end{itemize}

The experimental results (Table~\ref{tab:reduction_and_bon}) demonstrate that the Last scoring method consistently outperforms Mean, while BoN generally yields better results than Weighted BoN. This pattern holds across all four datasets and three TTS methods.

Notably, the superiority of Last over Mean suggests that some correct reasoning paths have high final scores but lower average scores. This indicates that even when reaching the correct answer, the intermediate reasoning steps may not be entirely accurate. Developing TTS methods that ensure both process and outcome accuracy remains a future research direction.

The choice between BoN and Weighted BoN reflects a balance between policy model and reward model decision-making. BoN relies primarily on PRM's judgment by selecting the highest-scoring path, while Weighted BoN considers the sampling frequency of the policy model through score aggregation. In our experiments, using a 1B parameter policy model and an 8B parameter PRM, the PRM-dominated BoN strategy achieved superior results, likely due to the PRM's stronger discriminative ability.

\subsection{Experiments on Qwen3-0.6B}
\label{app_sec:qwen3}

To validate the generalizability of SRCA, we conducted additional experiments on Qwen3-0.6B~\cite{yang2025qwen3technicalreport}, following the same settings as our main experiments. We set the sampling number N to 16 to expedite the experimental process. The experimental results are presented in Table~\ref{tab:qwen3}.
\begin{table}[ht]
\centering
\small
\setlength\tabcolsep{2pt}
\renewcommand{\arraystretch}{1.2}
\begin{tabular}{lcccc}
\hline
\multicolumn{1}{c}{\textbf{Models   \& TTS}} & \textbf{GSM8K}         & \textbf{MATH500}       & \textbf{AIME}        & \textbf{Olympiad} \\ \hline
Greedy Search                                & 42.61\%                & 34.40\%                & 3.54\%               & 13.63\%                \\
Self-Consistency                             & 52.62\%                & 47.00\%                & 4.07\%               & 20.59\%                \\
BoN                                          & 68.69\%                & 51.20\%                & 6.54\%                & 23.56\%                \\
Weighted BoN                                 & 63.91\%                & 53.60\%                & 7.29\%                & 23.41\%                \\
Beam Search                                  & 72.10\%                & 54.00\%                & 16.4\%                & 25.07\%                \\
DVTS                                         & 74.91\%                & 55.80\%                & 17.36\%                & 25.67\%                \\ \cdashline{1-5}
SRCA                                         & {\ul \textbf{79.45\%}} & {\ul \textbf{56.60\%}} & {\ul \textbf{21.33\%}} & {\ul \textbf{27.89\%}} \\ \hline
\end{tabular}
\caption{Comparative results of TTS with Qwen3-0.6B. Numerical values indicate accuracy rates, with bold figures denoting the best performance. Experimental parameters: N=16, utilizing DeepSeek PRM.}
\label{tab:qwen3}
\end{table}
Experimental results indicate that SRCA maintains superior performance compared to other TTS approaches. The observed trends align with the findings from our primary experiments, thereby confirming the general applicability of SRCA across different LLMs.

% \appendix

% \section{Example Appendix}
% \label{sec:appendix}

% This is an appendix.

\end{document}